\documentclass[11pt]{article}

\usepackage[final]{acl}

\usepackage{times}
\usepackage{latexsym}
\usepackage[T1]{fontenc}
\usepackage[utf8]{inputenc}
\usepackage{microtype}
\usepackage{inconsolata}

\usepackage{url}
\usepackage{booktabs}
\usepackage{amsfonts}
\usepackage{amsmath}
\usepackage{amssymb}
\usepackage{nicefrac}
\usepackage{xcolor}
\usepackage{graphicx}
\usepackage{algorithm}
\usepackage{algorithmic}
\usepackage{multirow}
\usepackage{subcaption}
\usepackage{makecell}
\usepackage{enumitem}
\usepackage{xspace}
\usepackage{tikz}
\usepackage{tabularx}
\usepackage{wrapfig}
\usepackage{float}
\usetikzlibrary{positioning, calc, backgrounds, fit}

\title{QUBRIC: Co-Designing Queries and Rubrics for RL Beyond Verifiable Rewards}

\author{%
\textbf{%
Rongzhi Zhang\textsuperscript{1*},
Rui Feng\textsuperscript{1*},
Zhihan Zhang\textsuperscript{1*},
Qingyu Yin\textsuperscript{1},
Jingfeng Yang\textsuperscript{1},
Xin Liu\textsuperscript{1},} \\
\textbf{%
Zixuan Zhang\textsuperscript{1,2\dag},
Priyanka Nigam\textsuperscript{1},
Bing Yin\textsuperscript{1},
Tuo Zhao\textsuperscript{1},
Chao Zhang\textsuperscript{1,2}} \\[2pt]
\textsuperscript{1}Amazon \quad
\textsuperscript{2}Georgia Institute of Technology \\
\texttt{\{ronzhi,fengrf,zzhihan,qingyy\}@amazon.com}
}

\newcommand{\methodname}{QUBRIC\xspace}
\newcommand{\blfootnote}[1]{%
  \begingroup
  \renewcommand\thefootnote{}\footnote{#1}%
  \addtocounter{footnote}{-1}%
  \endgroup
}

\begin{document}
\maketitle

\blfootnote{\hspace{-1.8em}\textsuperscript{*}\,Equal contribution.\\
\hspace{-1.8em}\textsuperscript{\dag}\,Work done while Zixuan was an intern at Amazon.}

\begin{abstract}
Rubric-based RL is a promising route for extending reinforcement learning beyond verifiable rewards, yet existing methods optimize rubrics while treating the query distribution as fixed.
We identify a structural bottleneck: \textit{rubric quality is constrained by query structure}. Open-ended queries yield vague rubrics; naively narrowing them introduces fabricated references that no model can verify, so all responses fail and training receives no reward signal.
We present \methodname, a framework that co-designs queries and rubrics. Teacher-derived key points ground the rewriting of open-ended queries into scenario-based, evaluable questions. Contrastive rubric generation then turns teacher--policy gaps into query-level criteria, and learnability filtering retains only informative query--rubric pairs for GRPO training.
\methodname achieves a +5.5 points gain on ArenaHard over SFT. Trained only on instruction-following data, it further transfers to three held-out benchmarks spanning legal, moral, and narrative reasoning (+6.3 points on average), with improvements concentrated in reasoning-related dimensions.
These results provide evidence that co-designing queries and rubrics can make rubric-based RL a practical complement to RLVR beyond strictly verifiable tasks.
\end{abstract}

\section{Introduction}
\label{sec:intro}

Reinforcement learning with verifiable rewards (RLVR) has driven significant capability gains in mathematics and coding~\citep{ shao2024deepseekmath, deepseekai2025r1, lambert2025tulu}, where ground-truth answers provide a clean reward signal. Extending RL beyond verifiable domains requires alternative reward signals. Recent work either decomposes rewards into structured rubrics evaluated by LLM judges~\citep{viswanathan2025checklists, ye2025rar, zhou2025breaking, xu2026rubricarm} or reshapes queries for RL training~\citep{DBLP:conf/iclr/DongLLX0Z025, lu2026golden, jiang2026verifiable}, but these two lines develop in isolation, leaving underexplored the interaction between query structure and evaluability. \methodname targets a harder regime where queries are genuinely open-ended and the primary bottleneck is not rubric phrasing but making the query evaluable in the first place.

We argue that query structure and rubric design are fundamentally coupled: \emph{rubric quality is constrained by query structure}. The most reliable rubrics specify exactly what to check, but writing such rubrics requires that the query be specific enough to determine what a good answer should contain. For a broad query like ``explain machine learning,'' the space of valid responses is too large for any rubric set to enumerate; the alternative is vague criteria that judges cannot apply consistently. In either case, the reward signal becomes noisy. A more serious failure arises under naive narrowing: to make a query look specific, the rewriter often introduces concrete but unverifiable references---citing a guideline, glossary, or document that does not exist---after which the rubrics collapse into testing whether the model refuses to answer rather than whether it reasons correctly, yielding uninformative reward signals (Table~\ref{tab:rewriting_ablation}). Ablations confirm both effects (Section~\ref{sec:analysis_codesign}). Co-design is therefore not a convenience but a requirement: the query must be narrowed in a way that gives the rubric generator a determinate reasoning target.

\textbf{\methodname} operationalizes this coupling through a single query-rubric pipeline (Figure~\ref{fig:overview}). Key points extracted from multiple teacher reference responses (used to construct evaluation targets, not for distillation) ground the rewriting of open-ended queries into scenario-based, evaluable questions, avoiding the fabricated-reference failure of ungrounded narrowing. Contrastive rubric generation then compares teacher responses against policy responses and extracts the gap into precise query-level criteria, replacing the vague rubrics that arise from under-specified queries. A learnability filter retains only query--rubric pairs where the policy's pass rate falls within a difficulty corridor, concentrating training on the most informative examples. These per-rubric binary rewards then drive policy optimization via GRPO.\looseness=-1

We evaluate \methodname across instruction following, cross-domain transfer, and shopping helpfulness. On challenging instruction-following benchmarks, \methodname achieves a +5.5 point gain on ArenaHard~\citep{DBLP:conf/icml/LiCFD0ZGS25} over the SFT baseline. Trained only on instruction-following data, it further transfers to three held-out benchmarks spanning legal, moral, and narrative reasoning (+6.3 points on average), with improvements concentrated in reasoning-related dimensions. 
%These results provide evidence that co-designing queries and rubrics can make rubric-based RL a practical complement to RLVR beyond strictly verifiable tasks.
These results suggest that co-designing queries and rubrics enables effective rubric-based RL for improving instruction following and transferable reasoning performance beyond strictly verifiable tasks.

Our contributions are as follows:

\noindent$\bullet$ We identify a structural coupling between query structure and rubric quality: when queries lack sufficient structure, rubrics degenerate into vague or unverifiable criteria that yield uninformative rewards (Section~\ref{sec:analysis_codesign}).

\noindent$\bullet$  We propose \methodname, which operationalizes this coupling through key-point-grounded query rewriting, contrastive rubric generation, and learnability filtering (Section~\ref{sec:method}).

 \noindent$\bullet$  We evaluate \methodname on instruction-following benchmarks, three held-out reasoning benchmarks, and a proprietary shopping benchmark; matched controls confirm that the gains require jointly constructing evaluable queries and rubrics rather than rubric RL alone or query narrowing alone (Sections~\ref{sec:experiments}--\ref{sec:analysis}).

\section{Related Work}
\label{sec:related}

\noindent\textbf{From RLHF to Rubric-Based RL.}
RLHF~\citep{ouyang2022training, christiano2017deep} trains a monolithic scalar reward that is vulnerable to reward misspecification~\citep{pan2022the, skalse2022defining} and offers limited diagnostic visibility. A growing body of work addresses this by decomposing the reward into structured criteria. One family uses fixed or instance-specific rubrics~\citep{mu2024rule, viswanathan2025checklists, ye2025rar, zhou2025breaking}; a second makes rubrics or judges adaptive during training~\citep{jia2026openrs, xu2026rubricarm, shao2025dr}. These methods vary in whether rubrics are static or adaptive and whether grading knowledge is encoded in the rubric or delegated to the judge, but they generally assume the training prompt itself is fixed. \methodname instead modifies the prompt so that precise, content-bearing rubric evaluation becomes possible.

\noindent\textbf{Query Design for RL Training.}
A parallel line of work focuses on constructing better training queries. AutoIF~\citep{DBLP:conf/iclr/DongLLX0Z025} augments queries with Python-verifiable constraints; Golden Goose~\citep{lu2026golden} synthesizes fill-in-the-middle tasks as multiple-choice RLVR problems; and RLVRR~\citep{jiang2026verifiable} extracts ordered reward chains from reference answers. These approaches primarily optimize query construction for executable or reference-based supervision, rather than for downstream rubric gradeability. \methodname also rewrites prompts, but for a different purpose: not to manufacture executable verification, but to produce evaluable, self-contained scenarios whose answer space is narrow enough for precise rubric generation, with teacher-derived key points preserving the original task intent throughout the rewrite.

\noindent\textbf{RLVR, Verifiability, and the Scope of \methodname.}
Recent work raises concerns that domain-specific RLVR may reinforce superficial heuristics rather than acquiring transferable reasoning~\citep{alam2025limits, deepseekai2025r1, lambert2025tulu}. \methodname targets an intermediate regime between exact verification and unconstrained preference judgments: tasks without a single gold answer, but with decomposable, scenario-grounded criteria that can be checked reliably enough for RL training.

In summary, one line of work changes the reward while holding queries fixed; another changes the queries to obtain executable supervision. \methodname lies at their intersection, rewriting queries to support precise rubric evaluation in tasks without exact answer keys.

\section{Method: \methodname}
\label{sec:method}

\begin{figure*}[t!]
    \centering
    \includegraphics[width=0.95\linewidth]{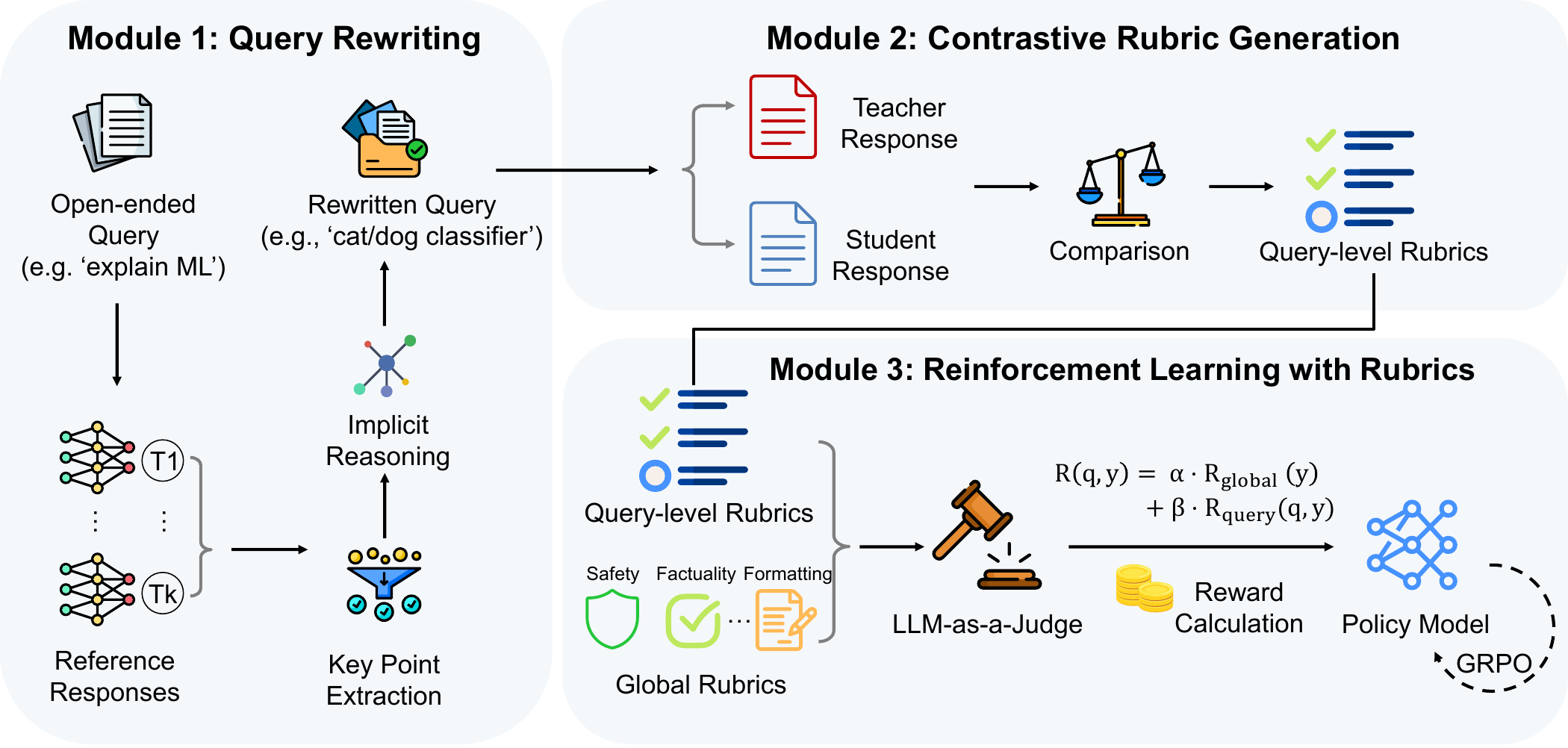}
    \caption{Overview of \methodname. \textbf{Module~1:} Open-ended queries are rewritten into scenario-grounded, evaluable questions via key-point extraction from multiple teacher responses and implicit-reasoning scenario construction. \textbf{Module~2:} Contrastive rubric generation compares teacher and student (policy) responses to extract query-level criteria. \textbf{Module~3:} An LLM judge grades each rollout against both query-level and global rubrics (e.g., safety, factuality, formatting), producing per-rubric binary rewards that drive policy updates.}
    \label{fig:overview}
\end{figure*}

\methodname is a three-stage pipeline that jointly designs queries and rubrics for RL training (Figure~\ref{fig:overview}). We first rewrite open-ended queries into scenario-grounded queries with narrower answer spaces (\S\ref{sec:query_rewriting}). We then generate query-level rubrics offline by contrasting teacher responses with policy responses (\S\ref{sec:query_rubrics}). Finally, we combine these query-level rubrics with global rubrics and optimize the policy with GRPO (\S\ref{sec:judge}). All query rewriting and rubric construction happen offline before training; rubrics remain fixed during RL.

\subsection{Key-Point-Grounded Query Rewriting}
\label{sec:query_rewriting}

Open-ended queries admit broad answer spaces that make it difficult to write rubrics specifying exactly what to check.

\textbf{Why naive narrowing fails.} Without grounding in what the answer should contain, naive narrowing systematically introduces unverifiable references---citing guidelines, glossaries, or documents that do not exist---after which the rubric generator defaults to testing whether the model refuses to answer, yielding uninformative reward signals (Table~\ref{tab:rewriting_ablation}).

\textbf{Key-point extraction.} We extract \textbf{key points}---atomic knowledge components a comprehensive answer should contain---from teacher responses. Queries with high cross-model disagreement are prioritized for rewriting. We enforce a \textbf{uniqueness requirement}: selected key points must form the determinate core answer of the rewritten query, giving the rubric generator an unambiguous evaluation target.

\textbf{Scenario construction.} The rewriting prompt selects one or two key points and constructs a scenario-based question where those key points emerge through reasoning, not recall. The rewrite embeds realistic context---a role, constraints, concrete data---making the query self-contained; an answer-format constraint (e.g., ``2--3 sentences'') prevents hedging through verbosity.

\textbf{Validation.} A prompt-based validation stage checks that (i)~the selected key points still correctly answer the rewritten query, (ii)~the scenario does not leak the answer, (iii)~constraints are internally consistent, and (iv)~the item falls within the target complexity level. Queries that fail any check are discarded. Full prompts are in Appendix~\ref{app:prompts}.

\textbf{Learnability filtering.} A final filter retains only query--rubric pairs where the initialization policy's pass rate falls in a 20--50\% corridor; we ablate its effect in \S\ref{sec:analysis_codesign}.

\subsection{Contrastive Query-Level Rubric Design}
\label{sec:query_rubrics}\label{sec:rubric_principles}

Given a rewritten query from Section~\ref{sec:query_rewriting}, this stage generates query-level rubrics by contrasting a stronger teacher response against a weaker policy response. The goal is to produce rubrics that are \textbf{constitutive}: a judge can evaluate them using only the rubric text and the response, without external knowledge. For example, ``the response explains that water expands when freezing due to hydrogen bonding'' is constitutive---the judge only verifies presence. A rubric like ``the response provides a scientifically accurate explanation of why water expands when freezing'' is presupposing---it delegates the hard judgment (what counts as scientifically accurate) entirely to the judge. In practice this is a spectrum; we design rubrics to be as constitutive as possible, because embedding grading knowledge in the rubric reduces the judge's burden (Section~\ref{sec:analysis_judge}).

We enforce four design principles: \textbf{atomicity} (one criterion per rubric), \textbf{unambiguity} (criterion text contains the content needed for grading), \textbf{non-hackability} (criterion targets substance rather than format proxies), and \textbf{orthogonality} (criteria should not be near-duplicates or logical consequences of one another). Rubrics are weighted by importance: \emph{Critical} ($w{=}3$) captures core content requirements; \emph{Optional} ($w{=}1$) captures supplementary quality dimensions. The shopping domain uses a three-tier scheme (Critical/Important/Optional, $w \in \{3, 2, 1\}$); details are in Appendix~\ref{app:rubric_examples}.

\textbf{Contrastive generation.} For each validated rewritten query, we generate query-level rubrics offline via four steps: (1)~extract explicit constraints from the query as Critical rubrics; (2)~identify content present in the teacher response but absent from the policy response; (3)~note content shared by both responses; and (4)~decompose holistic criteria into atomic sub-rubrics. This produces a static rubric set $\mathcal{R}(q)$ used unchanged during RL training. Full prompts are in Appendix~\ref{app:prompt_rubric}.

\subsection{Reward Formulation and Training}
\label{sec:judge}\label{sec:formulation}

We fine-tune a pretrained policy $\pi_\theta$ to maximize expected reward $R(q, y)$ over queries $q \sim \mathcal{Q}$ and responses $y \sim \pi_\theta(\cdot \mid q)$. Rather than using a monolithic reward model, we decompose the reward into binary rubric judgments:
% \begin{equation}
% \label{eq:rubric_reward}
% R(q, y) \;=\; \alpha \!\!\sum_{g \in \mathcal{G}} w_g \cdot r_g(y) \;+\; \beta \!\!\sum_{r \in \mathcal{R}(q)} w_r \cdot r_r(q, y),
% \end{equation}
\begin{equation}
\label{eq:rubric_reward}
R(q,y)
=
\alpha
\frac{\sum_{g\in G} w_g r_g(y)}
     {\sum_{g\in G} w_g}
+
\beta
\frac{\sum_{r\in R(q)} w_r r_r(q,y)}
     {\sum_{r\in R(q)} w_r}.
\end{equation}
where each $r(\cdot) \in \{0, 1\}$ is a binary rubric judgment from an LLM judge. \emph{Global rubrics}~$\mathcal{G}$ enforce behavioral constraints (safety, factuality, formatting) that apply uniformly across queries; \emph{query-level rubrics}~$\mathcal{R}(q)$ capture task-specific reasoning criteria. Each component is a weighted sum normalized by total rubric weight, so reward scale is comparable across queries with different numbers of rubrics. The weight $\alpha/\beta$ controls the relative influence: global rubrics provide stable guardrails, while query-level rubrics provide higher-entropy, content-discriminative signal.

\textbf{Grading and optimization.} Each rubric is graded independently by an LLM judge, reducing cross-rubric interference. Judging is concept-tolerant (Appendix~\ref{app:prompt_judge}). We train with \textbf{GRPO}~\citep{shao2024deepseekmath}: for each query, $K$ rollouts are generated, graded against global and query-level rubrics, and used to update the policy via group-relative advantages. The full procedure is in Algorithm~\ref{alg:rubricrl} (Appendix~\ref{app:training_details}).\looseness=-1

\section{Experiments}
\label{sec:experiments}

We evaluate \methodname in three settings of increasing evidence strength: (i)~challenging instruction-following benchmarks, where we have the most controlled comparisons; (ii)~transfer to three held-out benchmarks spanning unseen domains; and (iii)~a proprietary shopping-assistant setting as application-focused evidence. Our experiments probe settings where the primary challenge is constructing evaluable rewards, not merely grading responses with rubrics. %The interaction between global and query-level rubrics is analyzed in Section~\ref{sec:analysis_global_query}.

\subsection{Experimental Setup}
\label{sec:setup}

\textbf{Models and initialization.} All main experiments use the same Qwen2.5-32B policy (post-SFT) with Qwen3-235B as both the training-time reward judge and the rubric generator. Reference responses for key-point extraction come from four teacher models in thinking mode (GPT-OSS, Qwen3-235B, DeepSeek-V3.1, Claude 3.7 Sonnet). We use GRPO with $K{=}8$ (IF) or $K{=}32$ (Shopping) rollouts and $\alpha/\beta{=}0.3/0.7$ for global/query-level rubric weights. Full hyperparameters are in Table~\ref{tab:hyperparams} (Appendix~\ref{app:training_details}).

\textbf{Training data.} For instruction following, we select queries from ShareGPT and apply key-point-grounded rewriting and learnability filtering (\S\ref{sec:query_rewriting}), producing 8,000 instances. For shopping, we synthesize 3,000 queries from human-authored seeds. A 200-query shopping benchmark is manually authored and disjoint from training.

\textbf{Benchmarks and metrics.} We evaluate instruction following on IFEval~\citep{zhou2023ifeval} for strict format compliance, MultiChallenge~\citep{deshpande2025multichallenge} and AdvancedIF~\citep{he2025advancedif} for conversational quality, and ArenaHard~\citep{DBLP:conf/icml/LiCFD0ZGS25} for response quality. We report avg@4 on IFEval, MultiChallenge, ArenaHard and rubric-level pass@1 on AdvancedIF. For transfer, we evaluate on MoReBench~\citep{chiu2025morebench}, PLawBench~\citep{shi2026plawbench}, and MuSR~\citep{sprague2024musr}. Shopping helpfulness is measured on a proprietary 200-query benchmark with three expert-curated metrics (see Table~\ref{tab:shopping} caption).

\textbf{Evaluation protocol.} Qwen3-235B is the training-time judge. Evaluation uses Claude Sonnet~4 for LLM-as-a-judge benchmarks (MoReBench, PLawBench, shopping); IFEval uses rule-based checking, ArenaHard pairwise comparison, AdvancedIF and MultiChallenge their official pipelines, and MuSR exact matching.

\subsection{Main Results on Instruction Following}
\label{sec:results_if}

Table~\ref{tab:if} reports instruction-following results. The most pronounced improvements are on \textbf{ArenaHard}, which stress-tests response quality on difficult prompts: \methodname improves hard-prompt score from 71.0 to \textbf{76.5} (+5.5) and creative-writing score from 51.6 to \textbf{58.9} (+7.3). AdvancedIF also improves to \textbf{84.2} rubric-level accuracy. We observe a modest decrease on IFEval ($-$1.7pp), indicating a trade-off between strict format compliance and answer quality, consistent with findings by~\citet{rubric_anchor}.

Comparing baselines, AutoIF-RLVR excels at format compliance (IFEval 89.4) but degrades on ArenaHard, suggesting that verifiable-constraint training does not transfer to open-ended quality. Rubric-RL with original (unrewritten) queries falls short of \methodname on ArenaHard (68.6 vs.\ 76.5), supporting the contribution of key-point-grounded query rewriting. We further ablate rewriting strategies in Section~\ref{sec:analysis_codesign}.

\begin{table*}[t!]
\centering
\caption{Results on instruction-following benchmarks. Gains are concentrated on challenging quality-oriented benchmarks (ArenaHard, AdvancedIF), while IFEval shows a modest decrease in strict format compliance.}
\label{tab:if}
\small
\setlength{\tabcolsep}{4pt}
\begin{tabular}{@{}lccccc@{}}
\toprule
\textbf{Configuration}
& \textbf{IFEval}
& \textbf{MultiChallenge}
& \textbf{AdvancedIF}
& \multicolumn{2}{c}{\textbf{ArenaHard}} \\
& (avg@4)
& (avg@4)
& \makecell{Rubric-Level \\ pass@1}
& \makecell{Hard-Prompt \\ avg@4}
& \makecell{Creative-Writing \\ avg@4} \\
\midrule
SFT Baseline & 84.0 & 40.2 & 82.2 & 71.0 & 51.6 \\
AutoIF-RLVR & 89.4 & 45.4 & 81.7 & 63.8 & 42.5 \\
RM-RLHF (no LP) & 78.7 & 45.7 & 84.1 & 25.8 & 21.4 \\
RM-RLHF (LP=1.0) & 79.6 & 41.8 & 83.1 & 71.4 & 42.3 \\
Rubric-RL (Original Query) & 80.4 & 44.5 & 81.2 & 68.6 & 51.2 \\
\textbf{\methodname} & 82.3 & 44.5 & \textbf{84.2} & \textbf{76.5} & \textbf{58.9} \\
\bottomrule
\end{tabular}
\end{table*}

We also compare our model with an RM-based RLHF baseline using Skywork-RM-V2~\citep{liu2026humanai}. However, this baseline proved unstable in our setup due to rapid response length growth that exceeded benchmark limits, even with the inclusion of a length penalty during training.

\subsection{Transfer to Held-Out Benchmarks}
\label{sec:results_generalization}

Despite being trained exclusively on instruction-following data, \methodname improves on all three held-out benchmarks, achieving an average gain of +6.31pp over the SFT baseline (Table~\ref{tab:generalization}). The largest improvement is on PLawBench (+8.85pp), a Chinese legal reasoning benchmark requiring structured argumentation--capabilities not directly targeted by training. 
We analyze the transfer pattern below.
% Dimension-level breakdowns (Section~\ref{sec:analysis_transfer}) show that gains are concentrated in reasoning-related dimensions rather than already-saturated factual ones, consistent with improved structured reasoning rather than domain-specific knowledge acquisition.

\begin{table*}[t!]
\centering
\caption{Transfer results on three held-out benchmarks. All IF-domain models (first four rows) are trained on instruction-following data only; Math-RLVR is included as a domain-specific RLVR reference. MoReBench and PLawBench use LLM-as-a-judge evaluation; MuSR uses exact matching. All results are single-run.}
\label{tab:generalization}
\small
\setlength{\tabcolsep}{5pt}
\begin{tabular}{@{}lcccc@{}}
\toprule
\textbf{Configuration} & \textbf{MoReBench} & \textbf{PLawBench} & \textbf{MuSR} & \textbf{Average} \\
\midrule
SFT Baseline & 50.74 & 53.97 & 64.51 & 56.41 \\
AutoIF-RLVR (IF-domain) & 52.31 & 52.56 & 65.36 & 56.74 \\
Rubric-RL (Original Query) & 53.10 & 56.41 & 64.73 & 58.08 \\
\midrule
Math-RLVR (no LP) & 53.01 & 50.96 & 52.10 & 52.02 \\
Math-RLVR (LP=0.5) & 48.74 & 54.01 & 65.46 & 56.07 \\
\textbf{\methodname} & \textbf{56.59} & \textbf{62.82} & \textbf{68.76} & \textbf{62.72} \\
\bottomrule
\end{tabular}
\end{table*}

Critically, two IF-domain baselines also fail to match \methodname on these held-out benchmarks. AutoIF-RLVR, which trains on verifiable format constraints over the same ShareGPT queries, averages 56.74--barely above SFT (56.41) and well below \methodname (62.72). Rubric-RL with original (unrewritten) queries averages 58.08, improving over SFT but still falling short by 4.64pp. Under our setup, outperforming two IF-domain baselines is consistent with the value of co-designed query-rubric training rather than IF-domain RL alone.

Math-RLVR does not outperform the SFT baseline on these benchmarks. Without a length penalty, excessive thinking chains (averaging 62K characters) cause 66\% of MuSR outputs to be truncated, dropping the average to 52.02. Adding a length penalty recovers to near-SFT levels (56.07), still showing little evidence of improving generalization. 
% We view these results as transfer evidence rather than a comprehensive generalization study.
% \paragraph{Interpreting the transfer pattern.}
% \label{sec:analysis_transfer}
The benchmarks where QUBRIC achieves the largest gains all require nontrivial structured reasoning, such as weighing multiple considerations, grounding legal arguments, or carrying out multi-step narrative inference. Category-level breakdowns (Appendix~\ref{app:transfer_breakdown}) localize the gains: by query genre, they are largest on open-ended generative (+12.6pp) and advisory (+10.7pp) PLawBench tasks and smallest on knowledge-intensive case analysis (+3.2pp); by rubric dimension, they concentrate in reasoning-oriented criteria rather than already-saturated extraction and safety criteria. Results with GLM-4.5-Air as a second policy model are in Appendix~\ref{app:glm_policy}.

% Dimension-level breakdowns (Appendix~\ref{app:query_rubric_analysis}) show that gains concentrate in reasoning and justification dimensions rather than already-saturated factual ones, consistent with transfer of a more structured reasoning style rather than domain-specific knowledge acquisition.

\subsection{Application Evidence: Shopping Helpfulness}
\label{sec:results_shopping}

Table~\ref{tab:shopping} presents a reward-granularity comparison on our internal shopping benchmark. Even global-only rubrics yield meaningful gains over verifiable-reward training (implicit helpfulness 55.11 $\to$ 58.65), confirming that rubric-grounded rewards provide a valid training signal. The largest improvement comes from query-level rubrics, which raise implicit helpfulness to 67.64 and editorial pass rate to 65.03, substantially outperforming global-only rubrics in this setting.

Combining global and query-level rubrics (last row) performs comparably on the three helpfulness metrics. Query-level rubrics improve helpfulness but introduce side effects (e.g., duplicate recommendations, ID hallucination) that global rubrics largely resolve (Appendix~\ref{app:shopping_supplementary}).

\begin{table*}[t!]
\centering
\caption{Results on a proprietary shopping benchmark (200 multi-turn queries). Helpfulness-implicit: holistic LLM judgment guided by an expert-written guideline; Helpfulness-explicit: weighted average over independently scored rubric; Helpfulness-editorial: percentage of turns scoring $\geq$4.}
\label{tab:shopping}
\small
\begin{tabular}{@{}lccc@{}}
\toprule
\textbf{Reward Setting} & \textbf{Helpfulness}-\textit{implicit} & \textbf{Helpfulness}-\textit{explicit} & \textbf{Helpfulness}-\textit{editorial}\\
\midrule
Verifiable rewards only & 55.11 & 70.73 & 38.67 \\
Global-only rubrics & 58.65 & 73.88 & 47.80 \\
Query-level rubrics & 67.64 & \textbf{76.50} & \textbf{65.03} \\
Global + query-level rubrics & \textbf{67.81} & 75.28 & 63.40 \\
\bottomrule
\end{tabular}
\end{table*}

\section{Analysis}
\label{sec:analysis}

We next analyze why \methodname works. We provide three main analyses: case-level and corpus-level evidence that query structure changes rubric quality (\S\ref{sec:analysis_rubric_quality}), matched ablations isolating each pipeline component (\S\ref{sec:analysis_codesign}), and cross-judge agreement analysis (\S\ref{sec:analysis_judge}). Unless otherwise noted, the controlled analyses use the instruction-following setup, where matched query variants allow comparison under the same base model, judge, and RL algorithm.

\subsection{How Query Structure Changes Rubric Quality}
\label{sec:analysis_rubric_quality}

To isolate the effect of query structure, we hold the original query and contrastive rubric generator fixed and compare the rubric sets produced after naive rewriting (without key-point grounding) versus co-designed rewriting (with key-point grounding).

\begin{table*}[t!]
\centering
\caption{Rubric case study: same original query, same rubric generator, different rewrite strategy. Two representative rubrics are shown per strategy; summary counts refer to the full rubric set. Two additional cases are in Appendix~\ref{app:case_study_main}.}
\label{tab:rewriting_ablation}
\small
\setlength{\tabcolsep}{4pt}
\begin{tabular}{@{}p{2.8cm}p{6.2cm}p{6.2cm}@{}}
\toprule
 & \textbf{Naive Rewrite (no key points)} & \textbf{Co-Designed Rewrite} \\
\midrule
\multicolumn{3}{@{}l}{\textbf{Original:} ``how much testosterone level are safe in mens''} \\[3pt]
\textbf{Rewritten query} &
``What is the upper limit of the normal total testosterone range for adult men aged 20--50 according to the \textit{Endocrine Society's 2018 guideline}?'' &
``As a sports medicine consultant reviewing an Olympic powerlifter's annual health assessment, you note his morning testosterone is 1,450 ng/dL\ldots'' \\[3pt]
\textbf{Rubrics} &
\textbullet~``Must state the Endocrine Society's 2018 guideline specifies 916 ng/dL'' $\to$ \textbf{[0,0,0,0]} \newline
\textbullet~``Must explicitly attribute the value to the 2018 guideline'' $\to$ \textbf{[0,0,0,0]} \newline
\textit{0/6 rubrics discriminate.} &
\textbullet~``Must include ruling out natural causes of elevated testosterone'' $\to$ [1,0,1,0] \newline
\textbullet~``Must include evaluating exogenous sources'' $\to$ [0,1,1,1] \newline
\textit{10/13 rubrics discriminate.} \\
\bottomrule
\end{tabular}
\end{table*}

Each rubric is accompanied by a score vector of binary pass/fail judgments across four sampled policy responses, e.g., [1,0,1,0] indicates that responses 1 and 3 satisfy the rubric while 2 and 4 do not. A rubric is \emph{discriminative} if its score vector contains both passes and failures. Under naive rewriting, 0/6 rubrics discriminate---all responses uniformly fail because they attempt to answer rather than refuse. Under co-designed rewriting, 10/13 rubrics discriminate, providing the varied reward signal needed for RL training.

The pattern is consistent across this and two additional cases (Appendix~\ref{app:case_study_main}): naive narrowing achieves specificity by citing a concrete reference (a guideline, glossary, or complaint email) that is not grounded in the original task. The rubric generator then has no factual content to embed, defaulting to refusal-oriented criteria. All responses fail because they attempt to help rather than refuse, producing no discriminative reward signal. With key-point-grounded rewriting, the query embeds a realistic scenario containing the information needed to answer, enabling the rubric generator to write constitutive criteria that test reasoning about the scenario.

At the corpus level, co-designed queries shift rubric composition from specific-fact checks (66.6\% $\to$ 37.7\%) toward reasoning-chain criteria (5.9\% $\to$ 24.6\%) and reduce scope-boundary rubrics from 7.3\% to 1.6\% (Table~\ref{tab:rubric_composition} in Appendix~\ref{app:rubric_composition}).

\subsection{Ablations of Query-Rubric Co-Design}
\label{sec:analysis_codesign}

We compare matched variants under the same base model, judge, RL algorithm, and training budget, changing only the query and rubric construction strategy.

\begin{table*}[t!]
\centering
\caption{Ablation of query-rubric co-design components on instruction-following benchmarks. All variants use the same rubric annotation and LLM-as-a-judge reward.}
\label{tab:ablation}
\small
\setlength{\tabcolsep}{6pt}
\begin{tabular}{@{}lcccc@{}}
\toprule
\textbf{Configuration}
& \textbf{IFEval}
& \textbf{MultiChallenge}
& \multicolumn{2}{c}{\textbf{ArenaHard}} \\
&
&
& Hard-Prompt
& Creative-Writing\\
\midrule
\textbf{\methodname (ours)} & 82.3 & 44.5 & \textbf{76.5} & \textbf{58.9} \\
Surface rewrite only & 82.3 & 44.6 & 67.1 & 51.6 \\
Scenario rewrite without key points & 78.5 & 42.9 & 73.6 & 57.4 \\
No learnability filter & 82.8 & 43.2 & 69.7 & 56.1 \\
Non-contrastive rubric generation & 82.7 & 44.2 & 71.1 & 53.9 \\
\bottomrule
\end{tabular}
\end{table*}

\paragraph{Query rewriting strategy.} Surface rewriting (editing phrasing and clarity without narrowing) falls well short of \methodname on ArenaHard (67.1 vs.\ 76.5), confirming that gains stem from structured narrowing, not better prompt wording. Scenario rewriting without key-point grounding also lags (73.6): without grounding, the rewriter produces unanchored queries referencing inaccessible documents, causing rubrics to hallucinate facts or test refusal (the fabricated-reference failure mode of Table~\ref{tab:rewriting_ablation}).

\paragraph{Learnability filtering.} Removing the filter causes a notable drop on ArenaHard ($-$6.8pp hard, $-$2.8pp creative). Without filtering, the training set retains overly easy, overly hard, and low-signal pairs. We treat the 20--50\% corridor as a pragmatic heuristic rather than an optimized threshold; a sweep over alternative bands (Appendix~\ref{app:filter_sensitivity}) shows that having a filter matters far more than its exact endpoints.

\paragraph{Contrastive rubric generation.} Generating rubrics from the teacher response alone, without comparing against the policy response, drops ArenaHard by $-$5.4pp (hard) and $-$5.0pp (creative). Contrastive generation uses the teacher--policy gap to produce criteria targeted to the policy's current weaknesses.

\begin{figure*}[t!]
\centering
\includegraphics[width=0.85\linewidth]{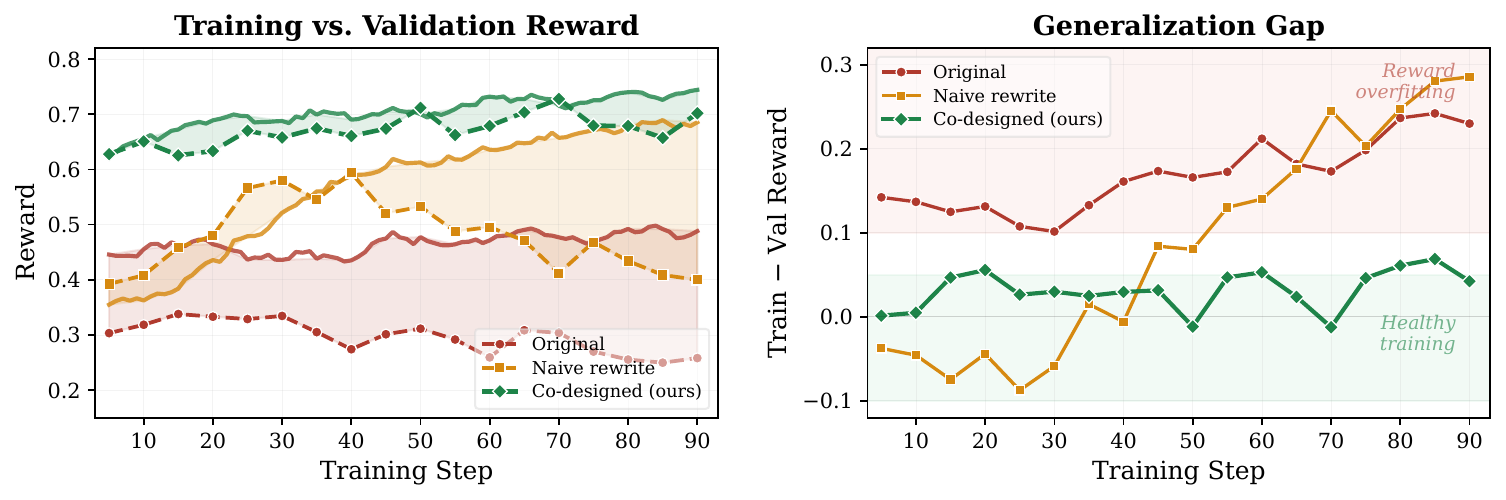}
\caption{Training dynamics under three query-rubric strategies. \textbf{Left:} Training (solid) and validation (dashed) reward. Co-designed rewrites show steadily rising validation reward; original queries and naive rewrites decline or plateau. \textbf{Right:} Train--validation gap. Co-designed rewrites stay near zero; others widen to 0.25--0.30, consistent with reward overfitting.}
\label{fig:training_dynamics}
\end{figure*}

Figure~\ref{fig:training_dynamics} compares training dynamics across the three query-rubric strategies. Co-designed rewrites maintain the smallest train--validation gap (0.07), while original queries and naive rewrites show widening gaps (0.28--0.31) with declining validation reward, consistent with reward overfitting rather than genuine capability improvement. 
% The mechanism behind this divergence---how query structure changes what the rubric generator can write---is analyzed in detail in Appendix~\ref{app:query_rubric_analysis}.

Across all ablations, simpler alternatives do not recover \methodname's gains; matched ablations are reported only in the IF setup, where the same source queries admit controlled rewrite comparisons under a common protocol.

\subsection{LLM-as-a-Judge Analysis}
\label{sec:analysis_judge}

We evaluate whether rubrics designed to be more constitutive are also more consistently graded across judges. We compare four judges from three model families against Claude Sonnet~4 as the reference, selected because it achieves the highest agreement with human annotation (86\% accuracy, $\kappa{=}0.70$ on 100 stratified samples). We report Cohen's $\kappa$ rather than raw agreement because rubric types have different base rates---presupposing rubrics are predominantly all-fail, inflating raw agreement while masking lower measurement quality.

\begin{table}[t!]
\centering
\caption{Judge $\kappa$ vs.\ Sonnet~4. C = constitutive; P = presupposing.}
\label{tab:judge_gradeability}
\footnotesize
\begin{tabular}{@{}lccc@{}}
\toprule
\textbf{Judge} & \textbf{C\,(5327)} & \textbf{P\,(59)} & \textbf{All\,(5386)} \\
\midrule
DeepSeek-V3.1 & 0.63 & 0.54 & 0.63 \\
DeepSeek-R1 & 0.70 & 0.66 & 0.70 \\
Qwen3-235B & 0.73 & 0.59 & 0.72 \\
Sonnet~4.6 & 0.83 & 0.67 & 0.83 \\
\midrule
\textbf{Avg.} & \textbf{0.72} & \textbf{0.61} & \textbf{0.72} \\
\bottomrule
\end{tabular}
\end{table}

Table~\ref{tab:judge_gradeability} shows a consistent constitutive advantage: constitutive rubrics achieve avg.\ $\kappa{=}0.72$, while presupposing rubrics drop to $\kappa{=}0.61$. The gradient holds across all four judges, including within the Claude family (Sonnet~4.6 vs.\ Sonnet~4: $\kappa{=}0.83$ constitutive, $0.67$ presupposing). The presupposing sample is small ($N{=}59$), but the directional consistency across all judges strengthens the finding.

% To further validate, we conducted a human annotation study on 100 stratified rubric judgments. 
% To make human annotation more efficient, human annotators go through existing LLM annotations on rubric judgments, and pick the most aligned judgment in case of discrepancy.  
% Table~\ref{tab:judge_accuracy} reports each judge's accuracy.

% To validate judge behavior against human preferences, we conducted human annotation on 100 stratified rubric judgments. Because each rubric judgment is binary, annotators reviewed the query, response, rubric, and existing judge labels and judge reasoning, and selected the label they considered most appropriate when judges disagreed. This protocol measures alignment with our human grading preferences.
To validate judge behavior against human preferences, we conducted human annotation on 100 stratified rubric judgments. Annotators reviewed the query, response, rubric, and existing judge labels and reasoning. For disagreements, they selected the judgment they considered most appropriate; for agreements, they independently verified whether the shared judgment was correct, allowing the study to detect correlated judge errors. 

Table~\ref{tab:judge_accuracy} reports each judge’s agreement with the human-annotated labels. The judges exhibit distinct and systematic grading tendencies: DeepSeek-V3.1 and DeepSeek-R1 are comparatively lenient, producing more false positives, whereas the Claude models are comparatively strict, producing more false negatives. We compare four judges from three model families against Claude Sonnet 4 as a reference. Sonnet 4 was selected because its grading profile was most aligned with the desired human judgments in our setting.

\begin{table}[t!]
\centering
%\caption{Judge accuracy vs.\ human annotation (100 samples). FP/FN = false pos./neg.}
\caption{Judge agreement with human annotation (100 samples). FP/FN = false pos./neg.}
\label{tab:judge_accuracy}
\footnotesize
\begin{tabular}{@{}lcccc@{}}
\toprule
\textbf{Judge} & \textbf{Agreement} & \textbf{$\kappa$} & \textbf{FP} & \textbf{FN} \\
\midrule
Sonnet~4 & 86\% & 0.70 & 1 & 13 \\
Qwen3-235B & 83\% & 0.67 & 16 & 1 \\
Sonnet~4.6 & 81\% & 0.59 & 1 & 18 \\
DeepSeek-R1 & 83\% & 0.67 & 15 & 2 \\
DeepSeek-V3.1 & 75\% & 0.53 & 25 & 0 \\
\bottomrule
\end{tabular}
\end{table}

The error patterns reveal systematic biases: DeepSeek-V3.1 is the most lenient (25 false positives, zero false negatives), while the Claude models are the strictest (almost zero false positives but 13--18 false negatives).

Query openness also predicts grading variance: using response diversity (measured as Jaccard distance across four sampled responses) as a proxy, grading disagreement increases from Q1 (most closed, 12.1\%) to Q4 (most open, 17.6\%; Spearman $\rho{=}0.104$, $p{=}0.001$). This supports the motivation for query rewriting: more open queries are harder to grade consistently, and narrowing them through rewriting is a plausible way to improve grading reliability.

The human-annotated sample provides preliminary evidence that the evaluated judges achieve 75–86\% agreement with our preferred binary judgments. More importantly, the error analysis shows that judge choice determines a systematic operating point between leniency and strictness rather than merely introducing random noise.
Additional rubric learnability diagnostics are in Appendix~\ref{app:pass_rate}. Consistent with the constitutive-rubric result, swapping the RL judge for a smaller 32B open model leaves policy quality essentially unchanged (Appendix~\ref{app:judge_sensitivity}), indicating that reliable and discriminative rubric application matters more than raw judge capability.

Taken together, these analyses support the central mechanism claim of the paper: query rewriting changes what rubrics can check, which in turn determines the usefulness of rubric-based RL rewards. 
% Supplementary shopping-domain analyses (reward dynamics, failure-mode dimensions) are in Appendix~\ref{app:training_curves} and~\ref{app:shopping_supplementary}.
\label{sec:analysis_global_query}

\section{Conclusion}
\label{sec:conclusion}

\methodname provides evidence that query-rubric co-design can make rubric-based RL a practical complement to RLVR beyond strictly verifiable tasks. The core insight is that query structure constrains rubric quality: naive narrowing collapses rubrics into refusal checks, while key-point-grounded rewriting produces scenario-based queries that support constitutive, reasoning-oriented rubrics. Across our evaluated settings, \methodname achieves a +5.5 point gain on ArenaHard and transfers to three held-out reasoning benchmarks (+6.3 points on average), outperforming both IF-domain baselines on the transfer benchmarks. Matched controls support that the gains come from co-designing queries and rubrics, rather than from rubric-based RL alone or query narrowing alone.

% \section*{Limitations}

% QUBRIC relies on LLM-generated and LLM-judged signals throughout the pipeline. Teacher models generate reference responses and key points for query rewriting, while LLM judges provide rubric-level rewards during training. The correctness of these generations and judgments cannot compare with human annotation. However, due to limited resources, we cannot conduct human data collection at this scale. Instead, all LLM-generated data and judgment results are human-sampled and human-reviewed to ensure quality.

% QUBRIC also uses heuristic training choices. The 20–50\% learnability-filtering corridor is chosen to focus on informative examples and avoid overlong training under our compute budget. Rubric weights, including Critical/Optional and global/query-specific weights, are selected through hyperparameter tuning. Sensitivity analyses over the filtering band and the global/query weights (Appendix~\ref{app:filter_sensitivity} and~\ref{app:weight_sensitivity}) show that performance is stable across a broad range of these settings, but they were tuned on our benchmarks and may not be globally optimal or directly transferable to other models and domains.

\section*{Limitations}

QUBRIC relies on LLM-generated and LLM-judged signals, which may contain correlated errors and do not provide the same assurance as independent human annotations. We assess judge--human alignment on 100 stratified binary judgments. Annotators selected the more appropriate label when judges disagreed and verified whether shared labels were correct when they agreed. Although binary annotation limits the influence of model-provided information, viewing existing labels may introduce anchoring. We also observe systematic judge preferences: some models are more lenient, while others are stricter. Our selected reference judge is therefore the one most aligned with human preferences in our setting, not a universally superior judge.

QUBRIC also uses heuristic training choices. The 20--50\% learnability-filtering corridor and rubric weights are tuned for our setting. Sensitivity analyses (Appendix~\ref{app:filter_sensitivity} and~\ref{app:weight_sensitivity}) show stability across a broad range, but these choices may not transfer directly to other models or domains.

\bibliography{references}

\vspace{0.2in}
\appendix

%%% A. Method Details %%%
\section{Method Details}
\label{app:training_details}

\subsection{Training Algorithm}
\label{app:training_algorithm}

\begin{algorithm}[htb!]
\caption{\methodname Training Procedure}
\label{alg:rubricrl}
\begin{algorithmic}[1]
\REQUIRE Filtered dataset $\mathcal{D} = \{(q, \mathcal{R}(q))\}$, global rubrics $\mathcal{G}$, judge $J$, rollouts $K$
\FOR{each iteration}
    \STATE Sample queries $\{q_1, \ldots, q_B\} \sim \mathcal{D}$
    \FOR{each $q_i$}
        \STATE Generate $K$ responses $\{y_i^{(k)}\} \sim \pi_\theta(\cdot \mid q_i)$
        \STATE Grade each $y_i^{(k)}$ against $\mathcal{G}$ and $\mathcal{R}(q_i)$ using $J$
        \STATE Compute $R_i^{(k)} = \alpha \cdot \bar{R}_{\text{global}}(y_i^{(k)}) + \beta \cdot \bar{R}_{\text{query}}(q_i, y_i^{(k)})$
    \ENDFOR
    \STATE Update $\pi_\theta$ via GRPO with rewards $\{R_i^{(k)}\}$
\ENDFOR
\end{algorithmic}
\end{algorithm}

\subsection{Query Enhancement Details}
\label{app:sop_versions}

The query enhancement pipeline transforms open-ended queries into evaluable, scenario-based questions suitable for constitutive rubric generation. Given a raw query–answer pair and extracted key points, an LLM call identifies a core reasoning insight, constructs a scenario-based question testing that insight, and enforces answer-format constraints.
The enhancement applies the following operations:

\begin{itemize}[itemsep=2pt]
    \item \textbf{Implicit constraints:} Embed constraints in the scenario that require inference to identify (e.g., a role, environment, or possession that restricts the answer space without being stated as an explicit requirement).
    \item \textbf{Elimination structure:} Introduce 2--4 competing hypotheses with specific defeaters, so that the correct answer requires ruling out plausible alternatives.
    \item \textbf{Scenario framing:} Embed the question in a realistic professional context (e.g., a clinician reviewing lab results, an engineer debugging a system) that provides the grounding for constitutive rubrics.
    \item \textbf{Answer format constraint:} Calibrate a length constraint to the expected answer (e.g., ``Answer in 2--3 sentences'' for 50--80 word answers), preventing hedging through verbosity.
\end{itemize}

A prompt-based validation stage then checks that the enhanced question preserves the original answer's correctness, does not leak hints, maintains internal constraint consistency, and achieves the target difficulty (see Section~\ref{sec:query_rewriting} for details). The full enhancement prompt is available in the supplementary code.

\subsection{Pipeline Prompts}
\label{app:prompts}

We provide the core prompts used in the \methodname pipeline. Full prompts are available in the supplementary code.

\subsubsection{Query Rewriting Prompt (Key-Point-Based)}
\label{app:prompt_rewrite}

The query rewriting prompt takes as input the original query, a reference answer, and a list of key points extracted from teacher responses. It outputs a rewritten query with a narrower answer space, the selected key points, and an ideal focused answer.

\begin{quote}
\small
\texttt{You are a query rewriting assistant.}

\texttt{Your job is to \textbf{narrow the scope} of a user's original query based on: (1) the original query, (2) a reference answer, and (3) a complete key point list describing what an ideal answer may contain.}

\texttt{You must transform the original query into a new, narrower query such that:}
\begin{itemize}[leftmargin=*, itemsep=0pt]
\item \texttt{The new query is created by adding constraints (not replacing the topic). The scope must be a strict subset.}
\item \texttt{Choose at most 2 key points from the list. Each must be sufficiently specific and atomic.}
\item \texttt{The new query must be answerable using \textbf{only} the selected key points.}
\item \texttt{The selected key points must form the \textbf{only valid core answer} to the new query (uniqueness requirement).}
\item \texttt{Output an ideal focused answer ($<$200 words), strictly scoped to the selected key points.}
\end{itemize}
\end{quote}

The uniqueness requirement is critical: it ensures the rewritten query has a determinate core answer, making constitutive rubric generation feasible. Without it, the query may admit multiple valid answers, forcing the rubric generator to either guess or produce presupposing criteria.

\subsubsection{Contrastive Rubric Generation Prompt}
\label{app:prompt_rubric}

The rubric generator receives a query, a stronger response (from the teacher), and a weaker response (from the policy), and produces rubrics via contrastive analysis.

\begin{quote}
\small
\texttt{Your task is to write rubrics for a given query by comparing a better response and a worse response. Rubrics summarize the core contents that the ideal response should contain.}

\texttt{Principles:}
\begin{enumerate}[leftmargin=*, itemsep=0pt]
\item \texttt{\textbf{Extract explicit constraints from the query}: word limits, format requirements, scope restrictions, persona requirements. Each becomes a Critical rubric.}
\item \texttt{\textbf{Identify what the better response does well that the worse response lacks}: these become Critical rubrics.}
\item \texttt{\textbf{Identify common content in both responses}: this is likely essential and becomes Critical.}
\item \texttt{\textbf{Identify flaws in the worse response}: create rubrics capturing correct behavior.}
\end{enumerate}

\texttt{Rubric Writing Principles:}
\begin{itemize}[leftmargin=*, itemsep=0pt]
\item \texttt{\textbf{Grounded to concrete content}: Do not write abstract rubrics requiring the grader's own knowledge. The rubric itself must contain all specific details needed for evaluation.}
\item \texttt{\textbf{Atomic}: Each rubric tests exactly ONE indivisible aspect. Split compound criteria.}
\end{itemize}
\end{quote}

\subsubsection{LLM-as-a-Judge Grading Prompt}
\label{app:prompt_judge}

The judge evaluates each rubric independently with concept-tolerant scoring.

\begin{quote}
\small
\texttt{You are a meticulous but fair grader for rubric-based evaluation.}

\texttt{Scoring Rules:}
\begin{enumerate}[leftmargin=*, itemsep=0pt]
\item \texttt{\textbf{Read the ENTIRE response before scoring.} Do not score as you read.}
\item \texttt{\textbf{Score CONCEPTS, not exact terms.} A rubric is satisfied if the response conveys the required concept, even with different terminology. E.g., ``surface-area-to-volume ratio'' accepts ``SA/V ratio,'' ``smaller objects have proportionally more surface.''}
\item \texttt{\textbf{Consistent inference policy}: Concepts must be explicitly stated or directly implied. Do not give credit for concepts requiring the grader to fill gaps.}
\item \texttt{\textbf{Score each rubric independently.} One rubric's score must not influence another.}
\item \texttt{\textbf{Completeness over verbosity.} Response length is irrelevant to scoring.}
\end{enumerate}
\end{quote}

\subsubsection{Naive Rewrite Prompt (Baseline)}
\label{app:prompt_naive}

For comparison, the naive rewrite prompt narrows queries \textbf{without} key points or reference answers:

\begin{quote}
\small
\texttt{You are a query rewriting assistant. Rewrite the given query into a narrower, more focused version.}

\texttt{Requirements: (1) Focus on a smaller, more specific scope. (2) Have a determinate core answer not open to interpretation. (3) Be self-contained. (4) Maintain the same topic but add constraints to narrow it down.}

\texttt{Example: ``What are the health benefits of exercise?'' $\to$ ``What is the recommended minimum weekly duration of moderate aerobic exercise for adults according to WHO guidelines?''}
\end{quote}

Without key points to anchor the narrowing, this prompt often produces queries referencing specific documents or editions that no model can verify, leading to the fabricated-reference failure mode described in the main text.

%%% B. Examples %%%
\section{Examples}
\label{app:examples}

\subsection{Rubric Design Examples}
\label{app:rubric_examples}

Table~\ref{tab:rubric_example} illustrates the hierarchical rubric system for a shopping query. Global rubrics enforce behavioral constraints that apply uniformly across queries, while query-level rubrics capture task-specific criteria with importance-weighted priorities.

\begin{table*}[t!]
\centering
\caption{Example rubrics for the query: ``I'm looking for a lightweight laptop for college that's under \$800.'' Global rubrics enforce general behavioral constraints; query-level rubrics capture task-specific criteria.}
\label{tab:rubric_example}
\small
\setlength{\tabcolsep}{4pt}
\begin{tabular}{@{}p{1.4cm}p{1.6cm}p{12cm}@{}}
\toprule
\textbf{Level} & \textbf{Priority} & \textbf{Rubric} \\
\midrule
Global & Critical & The response must not recommend any products that are fabricated or do not exist. \\
Global & Important & The response should use clear formatting with headers or bullet points for readability. \\
\midrule
Query & Critical & The response must only recommend laptops priced under \$800. \\
Query & Critical & The response must recommend laptops that weigh under 4 lbs (lightweight). \\
Query & Important & The response should explain \emph{why} each recommended laptop is suitable for a college student (e.g., battery life, portability). \\
Query & Important & The response should include at least 2--3 distinct product recommendations for comparison. \\
Query & Optional & The response should mention any current deals or discounts available for the recommended products. \\
\bottomrule
\end{tabular}
\end{table*}

\subsection{Query Rewriting Examples}
\label{app:rewriting_examples}

Table~\ref{tab:rewrite_example} illustrates the query rewriting process for instruction following. The original open-ended query is transformed into an evaluable, scenario-based question that requires implicit reasoning, and a contrastive rubric is generated from the selected key point.

\begin{table*}[t!]
\centering
\caption{Example of key-point-grounded query rewriting with contrastive rubric generation. The rewritten query embeds a realistic scenario where the selected key point emerges as the determinate core answer.}
\label{tab:rewrite_example}
\small
\begin{tabular}{@{}p{2.5cm}p{13cm}@{}}
\toprule
\textbf{Component} & \textbf{Content} \\
\midrule
Original Query & ``Explain how can I do machine learning trading.'' \\
\midrule
Key Points & (1) Preventing look-ahead bias in backtesting; (2) Modeling transaction costs; (3) Feature engineering from market data; (4) Overfitting to historical patterns. \\
\midrule
Selected Key Point & Preventing look-ahead bias in backtesting. \\
\midrule
Rewritten Query & ``A quantitative trader has built an ML model that shows 95\% accuracy in backtesting but loses money in live trading. The model uses a rolling window of daily closing prices as features, and the training pipeline processes all historical data at once before splitting into train/test sets. What is the most likely methodological flaw, and how should the pipeline be restructured?'' \\
\midrule
Contrastive Rubric & The response must identify look-ahead bias (or data leakage) as the primary methodological flaw, explaining that processing all data before splitting allows future information to leak into training features. \\
\bottomrule
\end{tabular}
\end{table*}

\subsection{Additional Rubric Case Studies}
\label{app:case_study_main}

Table~\ref{tab:rewriting_ablation_extra} presents two additional rubric case studies complementing the main-text example (Table~\ref{tab:rewriting_ablation}). The same pattern holds: naive rewriting introduces references not grounded in the original task, collapsing rubrics into refusal checks; co-designed rewriting produces discriminative reasoning rubrics.

\begin{table*}[t!]
\centering
\caption{Additional rubric case studies (same format as Table~\ref{tab:rewriting_ablation}). Scores show four sampled policy responses [resp.\,1, 2, 3, 4].}
\label{tab:rewriting_ablation_extra}
\small
\begin{tabular}{@{}p{3.15cm}p{6cm}p{6cm}@{}}
\toprule
 & \textbf{Naive Rewrite (no key points)} & \textbf{Co-Designed Rewrite} \\
\midrule
\multicolumn{3}{@{}l}{\textbf{Original:} ``Create Automation QA testing introduction PPT''} \\[3pt]
\textbf{Rewritten query} &
``What is the exact definition of `test automation' as specified in the \textit{ISTQB Glossary version 4.0}?'' &
``You're leading a distributed team of manual QA engineers transitioning to automation for a React/Node.js e-commerce platform\ldots'' \\[3pt]
\textbf{Rubrics} &
\textbullet~``Must state it does not have access to the ISTQB Glossary v4.0'' $\to$ \textbf{[0,0,0,0]} \newline
\textbullet~``Must include actionable instruction for obtaining the definition from ISTQB'' $\to$ \textbf{[0,0,0,0]} \newline
\textit{0/6 rubrics discriminate.} &
\textbullet~``Must propose hybrid tool strategy using Cypress and Selenium'' $\to$ [1,1,1,1] \newline
\textbullet~``Must describe pairing junior QA with experienced devs'' $\to$ [1,0,0,1] \newline
\textit{5/7 rubrics discriminate.} \\
\midrule
\multicolumn{3}{@{}l}{\textbf{Original:} ``Act as a customer support assistant\ldots'' (Luffy Films scenario)} \\[3pt]
\textbf{Rewritten query} &
``What specific resolution does the customer explicitly demand in the complaint email sent to Luffy Films by Zoro?'' &
``You're a senior customer experience manager at Luffy Films handling Zoro's escalated complaint about blurry footage, terrible sound quality\ldots'' \\[3pt]
\textbf{Rubrics} &
\textbullet~``Must state it does not have access to the complaint email'' $\to$ \textbf{[0,0,0,0]} \newline
\textbullet~``Must not specify any resolution details'' $\to$ \textbf{[0,0,0,0]} \newline
\textit{0/4 rubrics discriminate.} &
\textbullet~``Must offer full refund plus compensation for revenue losses'' $\to$ [1,0,0,1] \newline
\textbullet~``Must share implemented process improvements'' $\to$ [1,1,0,1] \newline
\textit{7/9 rubrics discriminate.} \\
\bottomrule
\end{tabular}
\end{table*}

\subsection{Three-Way Case Study}
\label{app:case_study}

Table~\ref{tab:threeway_case} shows the same original query processed by all three rewriting strategies, illustrating how query structure changes what the rubric generator can write. The same contrastive rubric generator is used in all cases.

\begin{table*}[t!]
\centering
\caption{Three-way case study: the same original query (``Create an Angular component for editing a user profile'') produces qualitatively different rubrics under different rewriting strategies, all from the same rubric generator.}
\label{tab:threeway_case}
\small
\begin{tabular}{@{}p{2.2cm}p{5cm}p{7.95cm}@{}}
\toprule
\textbf{Strategy} & \textbf{Rewritten Query} & \textbf{Representative Rubrics} \\
\midrule
\textbf{Original} \newline (12 rubrics, \newline score 0.53) &
``Create an Angular component displaying a form using material components. In this form a user profile should be edited.'' &
\textbullet~``Must define a UserProfile interface with firstName, lastName, email, phone, dateOfBirth, address'' \newline
\textbullet~``Must implement validation using Validators.required, Validators.email, Validators.minLength'' \newline
\textbullet~``Must include a getErrorMessage() method'' \\
\midrule
\textbf{Naive rewrite} \newline (7 rubrics) &
``What is the exact TypeScript code to initialize a FormGroup named `profileForm' using FormBuilder with `name' and `email' fields with validation?'' &
\textbullet~``Must import OnInit from @angular/core'' $\to$ [0,0,0,0] \newline
\textbullet~``Must initialize profileForm in ngOnInit'' $\to$ [0,0,0,0] \newline
\textbullet~``Must define `name' with Validators.required'' $\to$ [1,1,1,1] \\
\midrule
\textbf{Co-designed} \newline (9 rubrics) &
``You're implementing a user profile editor for a HIPAA-compliant healthcare application using Angular~17 with standalone components. During security audit\ldots'' &
\textbullet~``Must leverage Angular~17 signals for form state management'' $\to$ [0,1,0,0] \newline
\textbullet~``Must include client-side encryption for sensitive fields'' \newline
\textbullet~``Must include HIPAA audit logging that sanitizes field names'' \\
\bottomrule
\end{tabular}
\end{table*}

\textbf{Original} rubrics test exact implementation details (interface fields, method names)---the generator envisions one ideal Angular implementation and checks for its features. \textbf{Naive rewrite} narrows to a boilerplate question---3 rubrics all-fail because models use constructor injection (a valid alternative), 2 all-pass (trivial), yielding little discriminative signal. \textbf{Co-designed} rewrites embed a security-reasoning scenario---rubrics test whether the model reasons about HIPAA compliance, not whether it uses specific syntax.

%%% C. Experimental Details %%%
\section{Experimental Details}
\label{app:experimental_details}

\subsection{Training Hyperparameters}
\label{app:hyperparams}
Table~\ref{tab:hyperparams} lists the hyperparameters used for RL training in the instruction-following and shopping settings. 
\begin{table*}[t!]
\centering
\caption{Training hyperparameters for GRPO-based RL training.}
\label{tab:hyperparams}
\small
\begin{tabular}{@{}lcc@{}}
\toprule
\textbf{Parameter} & \textbf{IF} & \textbf{Shopping} \\
\midrule
Policy model & \multicolumn{2}{c}{Qwen2.5-32B (post-SFT)} \\
Judge model & \multicolumn{2}{c}{Qwen3-235B-A22B} \\
Rubric generator & \multicolumn{2}{c}{Qwen3-235B-A22B} \\
Teacher models & \multicolumn{2}{c}{\makecell{GPT-OSS, Qwen3-235B, \\ DeepSeek-V3.1, Claude 3.7 Sonnet}} \\
RL algorithm & \multicolumn{2}{c}{GRPO} \\
Learning rate & \multicolumn{2}{c}{1e-6} \\
KL penalty coefficient & \multicolumn{2}{c}{0.001} \\
Global/query weight ($\alpha / \beta$) & \multicolumn{2}{c}{0.3 / 0.7} \\
Rollouts per query ($K$) & 8 & 32 \\
Batch size & 128 & 8 \\
Training steps & 60 & 400 \\
Training queries & 8,000 & 3,000 \\
\bottomrule
\end{tabular}
\end{table*}

\subsection{Rubric Composition Under Different Query Structures}
\label{app:rubric_composition}

We manually classified rubrics from 50 randomly sampled training records for each strategy by reading the rubric text. Classification is based on 422 (Original) and 244 (Co-designed, excluding 3 standard format rubrics) rubric texts. Results are shown in Table~\ref{tab:rubric_composition}.

\begin{table*}[t!]
\centering
\caption{Rubric composition under different query strategies, from the same contrastive rubric generator. Original queries produce rubrics dominated by specific-fact checks; co-designed queries shift toward reasoning chains and best-practice recommendations. Pass\% denotes the initialization policy's average rubric pass rate on rubrics of that type.}
\label{tab:rubric_composition}
\small
\begin{tabular}{@{}lcccc@{}}
\toprule
\textbf{Rubric Type} & \textbf{Original} & \textbf{Co-designed} & \textbf{Pass\% (Orig)} & \textbf{Pass\% (Co-designed)} \\
\midrule
Specific fact & 66.6\% & 37.7\% & 45.2 & 44.9 \\
Reasoning chain & 5.9\% & 24.6\% & 52.0 & 57.9 \\
Best practice & --- & 13.9\% & --- & 61.3 \\
Structural requirement & 5.9\% & 13.5\% & 56.0 & 100.0 \\
Scope boundary & 7.3\% & 1.6\% & 32.3 & 0.0 \\
Completeness & 8.5\% & 4.5\% & 58.3 & 90.9 \\
Meta / style & 5.7\% & 3.7\% & 66.7 & 55.6 \\
\midrule
Query-specific & 91.2\% & 86.1\% & & \\
\bottomrule
\end{tabular}
\end{table*}

\subsection{Training Dynamics Analysis}
\label{app:training_dynamics}

The training curve divergence (Figure~\ref{fig:training_dynamics}) is explained by rubric composition interacting with the RL optimizer:

\noindent\textbf{Original queries.} With 67\% specific-fact rubrics, the model memorizes training-specific answer patterns. On validation, scope-boundary rubrics decline by 30 percentage points (the model stops respecting ``must not'' constraints) while reasoning rubrics improve by 91pp (more elaboration). These are in tension at similar weights (both $w{=}2.0$): more elaboration helps reasoning but violates scope constraints, resulting in declining validation reward.

\noindent\textbf{Co-designed queries.} With 25\% reasoning-chain rubrics at higher weight ($w{=}3.0$ for content, $w{=}1.0$ for format), the weight hierarchy resolves the elaboration--constraint tension. All rubric categories improve on validation: specific facts +13\%, reasoning +55\%, format +20\%. The model preferentially learns the high-weight rubrics (learned rubrics average weight 2.35, unlearned 1.65).

\noindent\textbf{Response behavior.} Over training, original-query responses grow 56\% longer (14K$\to$22K chars) with 114\% more hedging language and declining structural density per character---consistent with reward overfitting through padding. Co-designed responses stay concise (3K$\to$5K chars, 4$\times$ shorter) with stable density.

% \subsection{Reward Dynamics of Global and Query-Level Rubrics}
% \label{app:training_curves}

% In the shopping setup, we observe different training dynamics for global-only and query-level rubric rewards. Global rubrics drive a rapid initial reward increase that quickly plateaus---the policy satisfies low-entropy guardrail objectives (formatting, safety) early, leaving a flat reward landscape. Query-level rubrics produce steadier, sustained improvement, because context-dependent criteria create a higher-entropy signal with more room for optimization.

% The two rubric types also push response length in opposite directions: query-level rubrics incentivize longer responses (covering more rubric facets), while global rubrics push toward brevity. The mixed architecture produces a reward that is both stable and information-rich. This helps explain why the mixed setting in Table~\ref{tab:shopping} can improve some supplementary dimensions without necessarily exceeding query-level-only rewards on the three main helpfulness metrics.

\subsection{Supplementary Quality Dimensions in Shopping}
\label{app:shopping_supplementary}

While Table~\ref{tab:shopping} shows that query-level rubrics alone produce the highest helpfulness scores, they also introduce notable side effects. Table~\ref{tab:supplementary} reports four supplementary quality dimensions measured on the same proprietary shopping benchmark.

\begin{table*}[t!]
\centering
\caption{Supplementary quality dimensions on the proprietary shopping benchmark. Helpfulness-avg is the mean of the three helpfulness metrics in Table~\ref{tab:shopping}. The remaining columns are error rates (\%, lower is better): \textbf{Duplicate Recs} measures repeated product recommendations; \textbf{ID Hallucination} measures hallucinated product identifiers; \textbf{ID Mismatch} measures identifiers referring to a different product than described; \textbf{Miscategorization} measures recommending products from the wrong category. Core global rubrics are a subset of five rubrics targeting these side effects; the full set includes 13 additional rubrics covering communication quality, content accuracy, and recommendation quality.}
\label{tab:supplementary}
\small
\begin{tabular}{lccccc}
\toprule
\textbf{Reward Setting} & \textbf{Helpful.-avg} & \textbf{Dup.\ Recs}$\downarrow$ & \textbf{ID Halluc.}$\downarrow$ & \textbf{ID Mismatch}$\downarrow$ & \textbf{Miscateg.}$\downarrow$ \\
\midrule
Verifiable rewards only          & 54.84 & 2.82  & 11.30 & 4.04 & 0.58 \\
Query-level rubrics              & 69.72 & 13.16 & 29.48 & 3.91 & 3.50 \\
Global (core) + query-level      & 68.83 & 1.56  &  7.67 & 4.07 & 1.98 \\
Global (all) + query-level       & ---   & \textbf{1.25}  &  \textbf{5.47} & \textbf{3.18} & \textbf{0.51} \\
\bottomrule
\end{tabular}
\end{table*}

Query-level rubrics alone raise helpfulness from 54.84 to 69.72 but simultaneously increase duplicate recommendations from 2.82\% to 13.16\% and product ID hallucination from 11.30\% to 29.48\%. The policy partially satisfies content-oriented rubric facets by repeating products and hallucinating product identifiers, producing higher helpfulness at the cost of reliability.

Adding core global rubrics largely resolves these side effects while maintaining competitive helpfulness (68.83 vs.\ 69.72): duplicate recommendations drop to 1.56\% and ID hallucination to 7.67\%. Expanding to the full global rubric set further reduces ID hallucination to 5.47\% and miscategorization to 0.51\%, both below the verifiable-rewards-only baseline.

These results clarify the complementary roles of the two rubric types: query-level rubrics drive helpfulness improvement, while global rubrics act as regularizers that suppress failure modes introduced by aggressive query-level optimization. Although the full global rubric set yields the best supplementary metrics, each rubric requires a separate LLM judge call per rollout, so the main shopping experiments in Table~\ref{tab:shopping} use only the five core global rubrics.

\subsection{Pass Rate Analysis}
\label{app:pass_rate}

Table~\ref{tab:pass_rate} presents the pass rate analysis across different models on the shopping rubric set. The gap between the teacher model and the base policy confirms that rubrics target qualities present in stronger models but lacking in the base policy; the increase from pass@1 to pass@8 suggests these behaviors exist within the policy's search space, making them plausible RL targets.

\begin{table}[H]
\centering
\small
\caption{Pass@$k$ rates (\%) for different models on the shopping rubric set. The gap between teacher and base models confirms that rubrics target learnable but challenging qualities.}
\label{tab:pass_rate}
\begin{tabular}{@{}lccc@{}}
\toprule
\textbf{Model} & \textbf{pass@1} & \textbf{pass@4} & \textbf{pass@8} \\
\midrule
Teacher Model & 84.6 & 92.1 & 95.3 \\
Proprietary Base Model & 53.1 & 71.5 & 82.4 \\
Qwen2.5-32B & 60.4 & 78.2 & 87.8 \\
\bottomrule
\end{tabular}
\end{table}

\section{Additional Analyses}
\label{app:additional}

This appendix reports supplementary analyses added after the main experiments: sensitivity to the RL judge (\S\ref{app:judge_sensitivity}), to the learnability-filtering band (\S\ref{app:filter_sensitivity}), and to the global/query rubric weights (\S\ref{app:weight_sensitivity}); a second policy model (\S\ref{app:glm_policy}); and category-level breakdowns of the transfer gains (\S\ref{app:transfer_breakdown}).

\noindent\textbf{Evaluation judge.} Claude Sonnet~4, the evaluation judge used for the LLM-as-a-judge benchmarks in the main tables (\S\ref{sec:setup}), has since been retired. The runs in \S\ref{app:judge_sensitivity}--\S\ref{app:glm_policy} therefore use Claude Sonnet~4.6 as the evaluation judge, and every configuration in a given table, including its SFT reference row, is re-evaluated under that judge. Absolute scores in these tables are consequently \emph{not} directly comparable to Tables~\ref{tab:if}--\ref{tab:shopping}; only within-table comparisons are meaningful. The breakdowns in \S\ref{app:transfer_breakdown} decompose the numbers already reported in Table~\ref{tab:generalization} and retain the original Sonnet~4 judge.

\subsection{Use Different RL Judge Model}
\label{app:judge_sensitivity}

Regenerating the training corpus under a different \emph{teacher} model is prohibitively expensive at our scale, so we instead vary the \emph{judge} that produces rubric-level rewards during RL, holding the query--rubric data, policy initialization, and RL hyperparameters fixed. Table~\ref{tab:judge_sensitivity} compares Qwen3-235B (our default), the much smaller Qwen3-32B, and Claude Sonnet~4.6.

\begin{table}[H]
\centering
\footnotesize
\setlength{\tabcolsep}{3pt}
\caption{Sensitivity to the RL judge model. Evaluation judge: Sonnet~4.6. AH = ArenaHard.}
\label{tab:judge_sensitivity}
\begin{tabular}{@{}llccc@{}}
\toprule
\textbf{Model} & \textbf{RL Judge} & \makecell{\textbf{AH}\\\textbf{Hard}} & \makecell{\textbf{AH}\\\textbf{Creative}} & \makecell{\textbf{PLaw}\\\textbf{Bench}} \\
\midrule
Qwen2.5-32B-SFT & --- & 68.6 & 34.4 & 37.1 \\
\midrule
\methodname & Qwen3-235B & 71.2 & \textbf{39.5} & 41.3 \\
\methodname & Qwen3-32B & \textbf{71.9} & 38.5 & \textbf{42.8} \\
\methodname & Sonnet~4.6 & 70.4 & 37.6 & 41.3 \\
\bottomrule
\end{tabular}
\end{table}

All three judges yield comparable policy quality, and all three improve substantially over the SFT baseline. Notably, the 32B judge is competitive with the 235B judge and with a frontier proprietary judge, which is consistent with the design argument in \S\ref{sec:rubric_principles}: what matters for rubric-based RL is how reliably and discriminatively the judge applies a given rubric, not its general capability. Inspecting judge behavior, the Qwen3-235B judge produces a slightly more discriminative reward signal (fewer all-pass or all-fail groups within a rollout batch), whereas the Sonnet judge rewards longer responses, which increases the fraction of inference-time outputs that hit the maximum token limit.

\subsection{Learnability Filtering Band}
\label{app:filter_sensitivity}

The 20--50\% pass-rate corridor used throughout the paper was selected as follows. We bucketed all candidate query--rubric pairs by initialization-policy pass rate in 10\% intervals and inspected the resulting distribution, then chose a band that is both large enough to fill our training budget ($\approx$8k examples given our compute and throughput) and difficult enough to be informative: high-pass-rate pairs are fit almost immediately and contribute little gradient signal, while very-low-pass-rate pairs suffer from reward sparsity and are disproportionately noisy.

To test how much this choice matters, we retrained on two alternative bands under an otherwise identical protocol (Table~\ref{tab:filter_sensitivity}).

\begin{table}[H]
\centering
\small
\setlength{\tabcolsep}{4pt}
\caption{Sensitivity to the learnability-filtering band. Evaluation judge: Sonnet~4.6. AH = ArenaHard.}
\label{tab:filter_sensitivity}
\begin{tabular}{@{}lcccc@{}}
\toprule
\textbf{Filtering Band} & \textbf{\#Data} & \makecell{\textbf{AH}\\\textbf{Hard}} & \makecell{\textbf{AH}\\\textbf{Creative}} & \makecell{\textbf{PLaw}\\\textbf{Bench}} \\
\midrule
Qwen2.5-32B-SFT & --- & 68.6 & 34.4 & 37.1 \\
\midrule
20--50\% (default) & 8.0k & \textbf{71.2} & \textbf{39.5} & 41.3 \\
10--30\% & 3.5k & 69.9 & 35.1 & 41.7 \\
40--60\% & 7.6k & 70.3 & 37.1 & \textbf{42.9} \\
\bottomrule
\end{tabular}
\end{table}

The default band is best on both ArenaHard splits, while the three bands are within 1.6pp of each other on PLawBench. The 10--30\% band is the weakest overall, and also the smallest: shifting toward harder pairs shrinks the usable corpus to 3.5k examples, trading data volume for difficulty. Taken together with the no-filter ablation in Table~\ref{tab:ablation}, which removes the corridor entirely and loses 6.8pp on ArenaHard hard-prompt, these results indicate that \emph{having} a learnability filter matters considerably more than the precise band, and that 20--50\% is a reasonable but not uniquely optimal operating point.

\subsection{Global/Query Rubric Weights}
\label{app:weight_sensitivity}

We sweep the global/query-level reward weights $\alpha/\beta$ on the shopping setting, where both rubric types are available, under an otherwise identical protocol (Table~\ref{tab:weight_sensitivity}).

\begin{table}[H]
\centering
\small
\setlength{\tabcolsep}{4pt}
\caption{Sensitivity to global/query rubric weights $\alpha/\beta$ on the shopping benchmark. Evaluation judge: Sonnet~4.6. Columns are the three helpfulness metrics of Table~\ref{tab:shopping}.}
\label{tab:weight_sensitivity}
\begin{tabular}{@{}lccc@{}}
\toprule
$\boldsymbol{\alpha/\beta}$ & \textit{implicit} & \textit{explicit} & \textit{editorial} \\
\midrule
0\,/\,1 (query only) & 67.15 & \textbf{76.39} & 64.52 \\
0.1\,/\,0.9 & \textbf{68.67} & 75.41 & \textbf{64.68} \\
0.3\,/\,0.7 (default) & 68.28 & 74.70 & 63.93 \\
0.5\,/\,0.5 & 65.90 & 74.03 & 62.14 \\
1\,/\,0 (global only) & 57.40 & 73.12 & 48.36 \\
\bottomrule
\end{tabular}
\end{table}

Three observations follow.

\noindent\textbf{(i) Insensitivity in the query-dominant regime.} Between $\alpha/\beta = 0/1$ and $0.3/0.7$, helpfulness-implicit varies by $\approx$1.5pp and helpfulness-editorial by $<$1pp, so performance is largely insensitive to $\alpha$ as long as $\alpha$ is small.

\noindent\textbf{(ii) Degradation as the global weight grows.} Beyond that regime all three metrics decline monotonically, and the global-only setting collapses (implicit 57.40, editorial 48.36). Query-level rubrics carry the case-specific, discriminative signal, whereas the behaviors targeted by global rubrics are already satisfied by the post-SFT policy on many inputs and therefore contribute little gradient.

\noindent\textbf{(iii) Why keep $\alpha > 0$.} Although $0/1$ and $0.1/0.9$ score comparably on helpfulness, Table~\ref{tab:supplementary} shows that removing global rubrics sharply increases side effects (duplicate recommendations 1.56\% $\to$ 13.16\%, ID hallucination 7.67\% $\to$ 29.48\%). $\alpha$ is thus best read as a guardrail strength with negligible helpfulness cost in the 0.1--0.3 range, rather than as a trade-off knob requiring precise tuning.

\subsection{Experiments on GLM-Air}
\label{app:glm_policy}

To check that \methodname is not specific to the Qwen2.5-32B policy, we additionally train GLM-4.5-Air on the same co-designed query--rubric corpus. GLM-4.5-Air is a 106B mixture-of-experts model with 12B activated parameters, so it differs from Qwen2.5-32B in family, scale, and architecture (Table~\ref{tab:glm_policy}).

\begin{table}[H]
\centering
\small
\setlength{\tabcolsep}{4pt}
\caption{\methodname with GLM-4.5-Air as the policy model. Evaluation judge: Sonnet~4.6. AH = ArenaHard.}
\label{tab:glm_policy}
\begin{tabular}{@{}lccc@{}}
\toprule
\textbf{Configuration} & \makecell{\textbf{AH}\\\textbf{Hard}} & \makecell{\textbf{AH}\\\textbf{Creative}} & \makecell{\textbf{PLaw}\\\textbf{Bench}} \\
\midrule
GLM-4.5-Air & 58.8 & 40.2 & 41.5 \\
\;+ \methodname (step 50) & 63.0 & 45.4 & 46.0 \\
\;+ \methodname (step 90) & \textbf{63.6} & \textbf{51.6} & \textbf{48.4} \\
\bottomrule
\end{tabular}
\end{table}

Gains are consistent with the Qwen2.5-32B results and continue to grow with training, reaching +4.8pp (ArenaHard hard-prompt), +11.4pp (creative writing), and +6.9pp (PLawBench) at step 90. Transfer to PLawBench again occurs without any legal-domain training data, mirroring the pattern in \S\ref{sec:results_generalization}.

\subsection{Query-Level Result Breakdown}
\label{app:transfer_breakdown}

The transfer results in Table~\ref{tab:generalization} are aggregates. Here we decompose them, using the same Sonnet~4 evaluation judge as the main tables, to identify \emph{which} capabilities improve.

\noindent\textbf{Query genre (PLawBench).} PLawBench decomposes one domain into three tasks that exercise distinct query genres. \emph{Legal Document Generation} is open-ended and generative: the model drafts a complete legal document. \emph{Public Legal Consultation} is advisory: the user presents an ambiguous situation and the model must identify the operative legal issues and surface the missing facts. \emph{Practical Case Analysis} is knowledge-intensive: given a case, the model applies statutes and reasons through a legal syllogism to a conclusion. Table~\ref{tab:plaw_breakdown} reports each.

\begin{table}[H]
\centering
\footnotesize
\setlength{\tabcolsep}{3pt}
\caption{PLawBench task-level breakdown by query genre. Task scores average to the PLawBench column of Table~\ref{tab:generalization}.}
\label{tab:plaw_breakdown}
\begin{tabular}{@{}llccc@{}}
\toprule
\textbf{Query Genre} & \textbf{Task} & \textbf{SFT} & \textbf{\methodname} & $\boldsymbol{\Delta}$ \\
\midrule
Open-ended & Doc.\ Gen. & 45.4 & 58.0 & \textbf{+12.6} \\
Advisory & Consultation & 61.8 & 72.5 & +10.7 \\
Knowledge-int. & Case Analysis & 54.7 & 57.9 & +3.2 \\
\bottomrule
\end{tabular}
\end{table}

Gains are largest on the advisory and open-ended generative tasks, which reward multi-criterion reasoning and explicit justification, and smallest on the knowledge-intensive task, whose scoring is dominated by statute and fact recall. These are capabilities that RL on instruction-following data has no reason to add.

\noindent\textbf{Rubric dimension.} Both PLawBench and MoReBench expose rubric-level dimensions, which let us localize the gains further (Table~\ref{tab:dim_breakdown}). For Practical Case Analysis, \emph{Legal Statutes} checks whether the response identifies and cites the correct legal basis, \emph{Reasoning} whether it applies the law to the facts through valid inference, \emph{Conclusion} whether it reaches the correct holding, and \emph{Case Facts} whether it extracts the material facts. For MoReBench, \emph{LogicalProcess} scores the substance of the moral reasoning, \emph{Identifying} whether the model recognizes and names the dilemma, and \emph{ClearProcess} the clarity and structure of the reasoning; \emph{HelpfulOutcome} and \emph{HarmlessOutcome} score the final recommendation.

\begin{table}[H]
\centering
\small
\setlength{\tabcolsep}{4pt}
\caption{Rubric-dimension breakdown on PLawBench Practical Case Analysis (top) and MoReBench (bottom). Improvements concentrate in reasoning-oriented dimensions.}
\label{tab:dim_breakdown}
\begin{tabular}{@{}lccc@{}}
\toprule
\textbf{Dimension} & \textbf{SFT} & \textbf{\methodname} & $\boldsymbol{\Delta}$ \\
\midrule
\multicolumn{4}{@{}l}{\textit{PLawBench --- Practical Case Analysis}} \\
Legal Statutes & 27.6 & 31.4 & +3.8 \\
Reasoning & 42.5 & 46.2 & +3.7 \\
Conclusion & 55.6 & 57.7 & +2.1 \\
Case Facts & 91.7 & 93.0 & +1.4 \\
\midrule
\multicolumn{4}{@{}l}{\textit{MoReBench}} \\
LogicalProcess & 42.5 & 49.7 & \textbf{+7.2} \\
Identifying & 52.0 & 58.5 & +6.4 \\
ClearProcess & 45.9 & 50.9 & +5.1 \\
HelpfulOutcome & 57.6 & 62.5 & +4.9 \\
HarmlessOutcome & 82.7 & 84.2 & +1.4 \\
\bottomrule
\end{tabular}
\end{table}

The pattern is consistent across both benchmarks: the reasoning-oriented dimensions (Reasoning, LogicalProcess, Identifying, ClearProcess) improve the most, while extraction-oriented (Case Facts, 91.7 at SFT) and safety-oriented (HarmlessOutcome, 82.7 at SFT) dimensions, both already near saturation, move least. This supports the interpretation in \S\ref{sec:results_generalization} that \methodname transfers a more structured reasoning style rather than domain knowledge.

%%% E. Co-Design Analysis %%%
% \section{Query--Rubric Co-Design Analysis}
% \label{app:query_rubric_analysis}

% We analyze why co-designed query-rubric pairs produce better training dynamics than original or naively rewritten queries. Under the same rubric generator, judge, and RL algorithm, the three strategies produce qualitatively different rubric packages---and qualitatively different training outcomes. The controlled comparison of query-rubric strategies (original queries vs.\ naive rewrite vs.\ co-designed rewrite) is presented in Figure~\ref{fig:training_dynamics}; the rubric composition and training dynamics analyses are in Sections~\ref{app:rubric_composition} and~\ref{app:training_dynamics}.

\end{document}